\documentclass[sigconf]{acmart}
\AtBeginDocument{%
  }

\copyrightyear{2026}
\acmYear{2026}
\setcopyright{cc}
\setcctype{by}
\acmConference[MM '26]{Proceedings of the 34th ACM International Conference on Multimedia}{November 10--14, 2026}{Rio de Janeiro, Brazil}
\acmBooktitle{Proceedings of the 34th ACM International Conference on Multimedia (MM '26), November 10--14, 2026, Rio de Janeiro, Brazil}
\acmDOI{10.1145/3767308.3836182}
\acmISBN{979-8-4007-2213-4/2026/11}
\usepackage{enumitem}
\usepackage{placeins}  
\usepackage{tabularx}
\usepackage{algorithm}
\usepackage{algpseudocode}
\acmSubmissionID{7139}

\begin{document}

%
\title{Learning to Credit the Right Steps: Objective-aware Process Optimization for Visual Generation}

%
\author{Rui Li}
\authornote{Both authors contributed equally to this research.}
\email{rui.li@mail.ustc.edu.cn}
\affiliation{%
  \institution{University of Science and Technology of China}
  \city{HeFei}
  \country{China}
}
\affiliation{%
  \institution{Institute of Artificial Intelligence, China Telecom (TeleAI)}
  \city{Shanghai}
  \country{China}
}
\author{Ke Hao}
\authornotemark[1]
\email{ke\_hao2002@outlook.com}
\affiliation{%
  \institution{Shanghai Jiao Tong University}
  \city{ShangHai}
  \country{China}
}
\affiliation{%
  \institution{Institute of Artificial Intelligence, China Telecom (TeleAI)}
  \city{Shanghai}
  \country{China}
}

\author{Yuanzhi Liang}
\email{liangyzh18@outlook.com}
\affiliation{%
  \institution{Institute of Artificial Intelligence, China Telecom (TeleAI)}
  \city{Shanghai}
  \country{China}
}
\author{Chi Zhang}
\email{zhangc120@chinatelecom.cn}
\affiliation{%
  \institution{Institute of Artificial Intelligence, China Telecom (TeleAI)}
  \city{Shanghai}
  \country{China}
}

\author{Haibin Huang}
\email{jackiehuanghaibin@gmail.com}
\affiliation{%
  \institution{Institute of Artificial Intelligence, China Telecom (TeleAI)}
  \city{Shanghai}
  \country{China}
}

\author{Yun Gu}
\authornote{Corresponding authors.}
\email{geron762@sjtu.edu.cn}
\affiliation{%
  \institution{Shanghai Jiao Tong University}
  \city{ShangHai}
  \country{China}
}

\author{Xuelong Li}
\authornotemark[2]
\email{xuelong\_li@ieee.org
}
\affiliation{%
  \institution{Institute of Artificial Intelligence, China Telecom (TeleAI)}
  \city{Shanghai}
  \country{China}
}

\renewcommand{\shortauthors}{Rui Li et al.}

\begin{abstract}
Reinforcement learning, particularly Group Relative Policy Optimization (GRPO), has emerged as an effective framework for post-training visual generative models with human preference signals. However, its effectiveness is fundamentally limited by coarse reward credit assignment. In modern visual generation, multiple reward models are often used to capture heterogeneous objectives, such as visual quality, motion consistency, and text alignment. Existing GRPO pipelines typically collapse these rewards into a single static scalar and propagate it uniformly across the entire diffusion trajectory. This design ignores the stage-specific roles of different denoising steps and produces mistimed or incompatible optimization signals.
To address this issue, we propose Objective-aware Trajectory Credit Assignment (OTCA), a structured framework for fine-grained GRPO training. OTCA consists of two key components. Trajectory-Level Credit Decomposition estimates the relative importance of different denoising steps. Multi-Objective Credit Allocation adaptively weights and combines multiple reward signals throughout the denoising process. By jointly modeling temporal credit and objective-level credit, OTCA converts coarse reward supervision into a structured, timestep-aware training signal that better matches the iterative nature of diffusion-based generation. Extensive experiments show that OTCA consistently improves both image and video generation quality across evaluation metrics.
\end{abstract}

%
%




\maketitle

\section{Introduction}
\begin{figure}[t]
    \centering
    \includegraphics[width=1\linewidth]{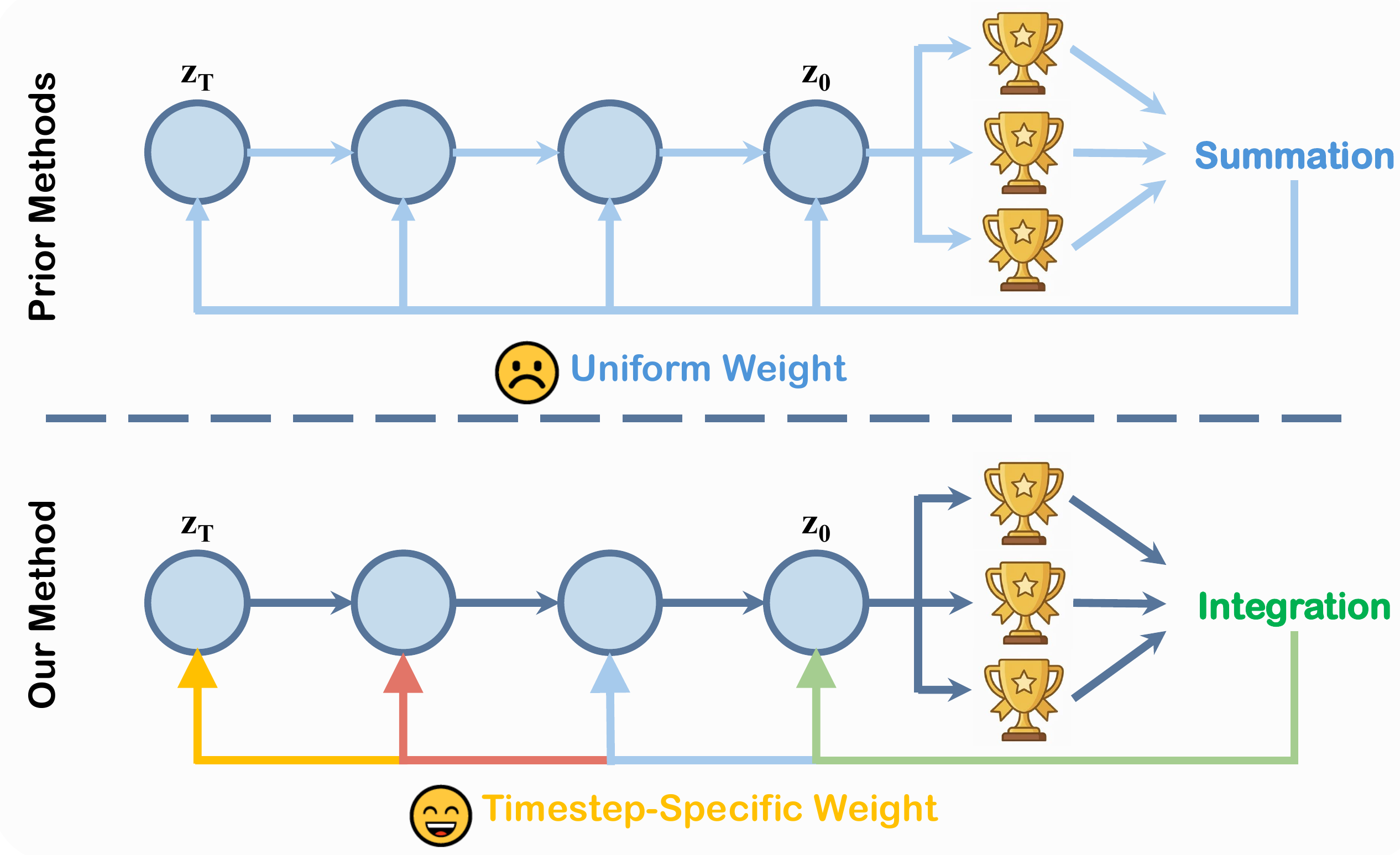}
    \caption{Comparison between our framework and existing GRPO-based approaches.}
    \label{fig:placeholder}
\end{figure}
Reinforcement learning (RL) is emerging as an effective paradigm for post-training visual generative models with human preference signals~\cite{diffusion_ho,diffusion_song,flow_matching,rectifiedflow,firstrlddpm,fan2023dpok}. Among existing methods, Group Relative Policy Optimization (GRPO)~\cite{deepseek} is particularly attractive as its group-wise relative comparisons stabilize updates and improve sample quality. Recent adaptations to diffusion and flow-based generation~\cite{dancegrpo,liu2026flow} further position GRPO as a promising framework for reward-driven visual alignment~\cite{an2026ai,liang2026integrating,zhang2024vast,liang2026teleboost,liu2026learning,li2025rethinking,liang2022seeg,liang2022simple,li2022positive,zhang2025variational,chen2026generative}.

However, the effectiveness of visual GRPO is heavily shaped by the quality and design of reward supervision. Unlike supervised learning with explicit targets, GRPO improves the policy through reward-driven relative updates~\cite{mao2025image,zheng2026manifold}. As a result, training quality depends directly on whether the reward signal is informative, compatible, and well organized. In practice, visual generation rarely admits a single sufficient reward. Modern pipelines therefore combine multiple reward models to capture different objectives, such as visual quality, motion coherence, and text alignment\cite{vader,dufour2025miro}. But these rewards are often not aligned: different reward models may favor different outputs, and naïvely combining them can produce conflicting optimization directions\cite{zhang2024learning,visionreward,lee2025calibrated}.

However, in GRPO-style SDE sampling, different timesteps do not contribute equally to the final output~\cite{zhou2025g2rpo,zhang2026grpo,vu2026memorization}. Existing GRPO pipelines largely ignore this heterogeneous step-wise contribution. They broadcast the same reward supervision uniformly across all timesteps, as if every generation step were equally responsible for the final outcome. This leads to misassigned policy credit: beneficial steps may be under-emphasized, while unhelpful or even detrimental steps may receive the same reinforcement, ultimately producing poorly coordinated updates along the trajectory.

We observe that the key challenge in visual GRPO is not only how to optimize, but how to organize multi-reward supervision into compatible and timestep-appropriate policy updates. To this end, we propose Objective-aware Trajectory Credit Assignment(OTCA), a structured reward credit modeling framework that performs timestep-specific credit assignment over heterogeneous reward objectives throughout the flow trajectory. OTCA combines two complementary components for temporal credit modeling and objective coordination, allowing each generation step to receive both an appropriate optimization strength and a timestep-aware objective mixture. In this way, steps more causally important to the final output receive stronger policy updates, while conflicting rewards are adaptively reconciled according to their relevance at that stage of generation. As a result, OTCA transforms multi-reward supervision from a static global signal into a structured, timestep-aware training signal. Our contributions can be summarized as follows:

\begin{itemize}[noitemsep, topsep=0pt, parsep=0pt, partopsep=0pt, leftmargin=1em]
    \item We propose \textbf{Objective-aware Trajectory Credit Assignment (OTCA)}, a structured reward credit modeling framework that refines the coarse global supervision into timestep-level policy updates. OTCA jointly determines both the optimization strength and the objective mixture over heterogeneous reward signals at each timestep. 
    
    \item We introduce Multi-Objective Credit Allocation (MOCA) to integrate multiple reward signals in the advantage space, avoiding optimization in the high-dimensional gradient space. We further propose Trajectory-Level Credit Decomposition (TCD) to estimate timestep-wise contribution under SDE sampling. TCD directly leverages the sampling trajectories produced by GRPO and introduces no extra modules. 

    \item Extensive experiments show that OTCA consistently improves optimization stability and visual alignment quality, yielding superior performance across multiple metrics.
    
\end{itemize}

\section{Related Work}

\label{sec:related}
\noindent\textbf{RL for Visual Generation.}
Inspired by Proximal Policy Optimization (PPO)~\cite{ppo}, early works~\cite{firstrlddpm,ddpo,fan2023dpok} integrated reinforcement learning into diffusion models by optimizing the score function~\cite{song2020score} with policy gradient methods, thereby enabling image generation that better aligns with human preferences.
More recently, GRPO-based approaches~\cite{dancegrpo,liu2026flow,mixgrpo,tempflow,li2025growing,ni2025seeing} have advanced visual generation to new heights. In particular, DanceGRPO~\cite{dancegrpo} and FlowGRPO~\cite{liu2026flow} adapt GRPO to visual generation by reformulating Flow Matching’s~\cite{flow_matching} ODE sampling into an SDE formulation, enabling online RL training on state-of-the-art generative models. VIPO~\cite{ni2025seeing} further enhances optimization by focusing on salient video regions through visual feature extraction, while Li et al.~\cite{li2025growing} leverage large-scale pretrained VLMs to jointly grow the generator and reward model.

Despite their success, these methods still apply a single scalar advantage uniformly across all timesteps, overlooking the heterogeneous contributions of different denoising steps. This coarse treatment ignores the nuanced information embedded in each timestep, limiting effective credit assignment in reinforcement-driven visual generation.
To alleviate this issue, several recent works attempt to mitigate credit assignment ambiguity in flow-based GRPO through modified sampling strategies~\cite{zhou2025g2rpo,zhang2026grpo,ge2025expand,he2025neighbor}.
Among them, G²RPO~\cite{zhou2025g2rpo} explicitly identifies the sparse reward attribution problem in flow-based GRPO and introduces singular stochastic sampling. By restricting stochasticity to individual denoising steps, G²RPO seeks to provide more precise step-wise credit signals.
E-GRPO~\cite{zhang2026grpo} argues that high-entropy denoising steps drive effective exploration. It merges low-entropy steps and restricts stochastic updates to high-entropy steps to reduce reward ambiguity.
While these approaches provide practical improvements, they operate by modifying how trajectories are sampled rather than changing how rewards are attributed. The supervision signal remains defined at the trajectory level, and its propagation through multi-step stochastic transitions is left intact. Consequently, the core mismatch between trajectory-level rewards and step-wise updates persists.

\noindent\textbf{Reward Modeling and Multi-objective Optimization.}
Applying reinforcement learning to visual generation critically depends on reliable reward modeling. In image generation, preference-based models such as PickScore~\cite{pickscore}, HPSv2~\cite{hpsv2}, and ImageReward~\cite{xu2023imagereward} estimate human visual preferences directly in pixel space. For video generation, VideoScore~\cite{he2024videoscore} and VideoAlign~\cite{videoalign} extend this paradigm by evaluating visual quality, motion consistency, and text alignment, while VisionReward~\cite{visionreward} introduces more fine-grained visual feedback signals. 

Despite their effectiveness, these reward models operate on final rendered outputs and provide only global scalar supervision. They are inherently trajectory-agnostic and cannot offer timestep-level feedback within the diffusion process, where intermediate latent states determine the final outcome. In practice, post-training for visual generation typically relies on multiple reward models to jointly optimize different evaluation dimensions. This naturally formulates the problem as multi-objective optimization (MOO). In such settings, multiple objectives may exhibit conflicting optimization directions, and fixed reward weighting fails to adaptively balance the trade-offs among them. Classical gradient-based MOO methods attempt to resolve these conflicts by explicitly operating on per-objective gradients, e.g., MGDA~\cite{desideri2012multiple} computes a minimum-norm convex combination of task gradients, while PCGrad~\cite{yu2020gradient} and CAGrad~\cite{liu2021conflict} modify conflicting gradients to reduce interference. However, these approaches require computing and storing separate gradients for each objective at every update step, which becomes infeasible for large-scale diffusion models.
To alleviate this limitation, MGDA-UB~\cite{sener2018multi} derives an upper bound of the multi-objective gradient norm and proves that minimizing this bound leads to a Pareto-stationary solution under mild assumptions. By optimizing in a scaled representation space, it avoids explicit per-objective gradient computation.
Another line of work addresses objective imbalance through adaptive loss reweighting methods~\cite{chen2018gradnorm}, which dynamically adjust task weights based on training dynamics. However, diffusion-based alignment involves rapidly evolving objectives over limited denoising steps, rendering such dynamic estimation unreliable.
More recently, Multi-GRPO~\cite{lyu2025multi} addresses objective imbalance in a post-training RL setting by normalizing each reward independently. This strategy primarily rescales reward magnitudes and provides limited support for true multi-objective optimization.

\section{Method}
\label{sec:method}

We propose OTCA, a structured reward credit modeling framework for GRPO-based visual generation post-training. 
Our central premise is that effective policy optimization requires assigning appropriate reward signals to the right generation timesteps.
To achieve this, OTCA performs joint credit assignment over timesteps and reward objectives within a unified framework, so that each generation step is assigned a credit signal that simultaneously reflects its trajectory contribution and the appropriate reward composition for optimization.
Formally, the effective advantage at timestep $t$ for sample $i$ is defined as
\[
\tilde{A}_{t}^{i} 
= 
w_t^{i} \cdot 
 \sum_{k} c_{k}^{i} A_k^{i},
\]
where $w_t^{i}$ reflects the relative contribution of timestep $t$ to the final outcome, and $c_{k}^{i}$ specifies how different reward signals should be composed at that step. Together, they define a timestep-conditioned credit signal that determines the supervision received by each denoising step, allowing policy updates to follow the actual structure of the generation trajectory rather than relying on a single global reward broadcast.
\begin{figure*}
    \centering
    \includegraphics[width=0.9\linewidth]{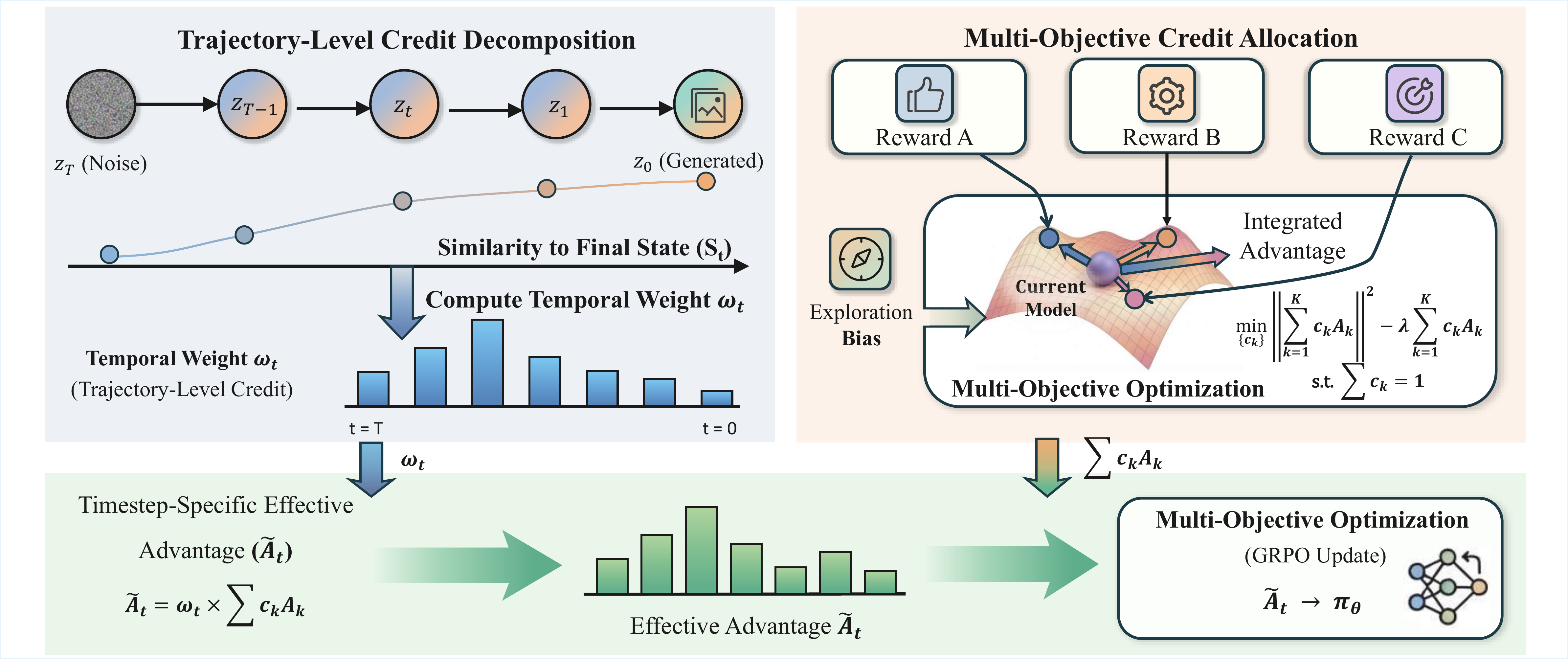}
\caption{Overview of our method. Different from existing GRPO-based approaches, our pipeline explicitly injects multiple objective optimization signals into each temporal timestep dimension, enabling balanced optimization at the timestep level.}

    \label{fig:placeholder}
\end{figure*}
\subsection{Preliminaries}
\label{sec:pre}

\noindent\textbf{GRPO for Visual Generation.}
The iterative denoising procedure in diffusion and flow-based models can be viewed as a sequential decision process, where each timestep corresponds to a policy action applied to the current latent state. Following \cite{deepseek}, we adopt Group Relative Policy Optimization (GRPO) to optimize the generative policy $\pi_{\theta}$.
Given a group of $G$ samples generated from the same prompt, GRPO maximizes the following clipped objective:

\begin{multline} \label{eq:grpo} \mathcal{J}(\theta) = \mathbb{E}_{\{\mathbf{o}_i\}\sim\pi_{\theta_{\text{old}}}}\Bigg[\frac{1}{G}\sum_{i=1}^{G}\frac{1}{T}\sum_{t=1}^{T} \min\Big(\rho^i_{t} A_i,\\ \text{clip}\left(\rho^i_{t}, 1-\epsilon, 1+\epsilon\right) A_i\Big)\Bigg], 
\end{multline}

where $\rho^i_{t} = \frac{\pi_{\theta}(\mathbf{a}_{t,i}\mid \mathbf{s}_{t,i})}{\pi_{\theta_{\text{old}}}(\mathbf{a}_{t,i}\mid \mathbf{s}_{t,i})}$ denotes the importance ratio between the current and previous policies, and $\pi_{\theta}(\mathbf{a}_{t,i}\mid \mathbf{s}_{t,i})$ is the policy probability for action $\mathbf{a}_{t,i}$ at state $\mathbf{s}_{t,i}$.

The advantage $A_i$ is computed using group-wise normalized rewards:

\begin{equation}
\label{eq:advantage}
A_i = \frac{r_i - \text{mean}({r_1, r_2, \dots, r_G})}{\text{std}({r_1, r_2, \dots, r_G})}.
\end{equation}

This normalization measures the relative performance of each sample within the group and stabilizes policy optimization.

\noindent\textbf{Stochastic Sampling for Flow Models.}
Flow-matching models typically define generation through a deterministic ordinary differential equation (ODE),
\begin{equation}
\mathrm{d}\mathbf{z}_t = \mathbf{u}_t\,\mathrm{d}t,
\end{equation}
where $\mathbf{u}_t$ denotes the learned velocity field. While this formulation is effective for generation, its deterministic nature produces only a single trajectory for each initialization, which limits the trajectory diversity required by policy optimization methods such as GRPO.

To introduce exploration, we follow prior work and consider a reverse-time stochastic differential equation (SDE),
\begin{equation}
\mathrm{d}\mathbf{z}_t
=
\left(
\mathbf{u}_t
-
\frac{1}{2}\varepsilon_t^2 \nabla_{\mathbf z_t}\log p_t(\mathbf{z}_t)
\right)\mathrm{d}t
+
\varepsilon_t\,\mathrm{d}\mathbf{w},
\end{equation}
where $\varepsilon_t$ controls the noise magnitude and $\mathrm{d}\mathbf{w}$ is standard Brownian motion. Compared with the deterministic ODE, this SDE induces stochastic trajectory sampling and thus enables exploration over multiple denoising paths. The score function can be written in closed form as
\begin{equation}
\nabla_{\mathbf z_t}\log p_t(\mathbf{z}_t)
=
-\frac{\mathbf{z}_t-\alpha_t\mathbf{x}}{\sigma_t^2}.
\end{equation}
Substituting this score into the reverse-time SDE yields a stochastic sampling process that can be naturally interpreted as a policy $\pi_\theta(\mathbf a_t\mid \mathbf s_t)$, making flow-based generation compatible with GRPO-style reinforcement learning.

\subsection{Trajectory-Level Credit Decomposition}
\label{sec:temporal}

In GRPO, the trajectory-level advantage $A_i$ is uniformly broadcast to all timesteps, implicitly treating each denoising step as equally credit-worthy. This assumption can be overly coarse for diffusion-based generation, where different timesteps may contribute unevenly to the final outcome. The issue can become more pronounced under SDE sampling~\cite{song2020score}, where stochastic exploration introduces additional variability into the denoising trajectory. As a result, some intermediate transitions may provide clearer evidence of useful progress, while others may be less informative or temporarily less aligned with the eventual denoised result. Assigning the same policy signal to all such steps can therefore introduce unnecessary variance into optimization and yield unreliable updates.

To make policy optimization more sensitive to the denoising trajectory, we introduce a temporal credit assignment mechanism that estimates the relative importance of each timestep. Inspired by prior work that uses cosine similarity as a scale-invariant measure of representation alignment~\cite{chen2020simple,grill2020bootstrap}, we quantify how well each intermediate latent state aligns with the final denoised latent state~\cite{pang2025aligning}. Let $z_t^{i}$ denote the latent representation of the $i$-th sample at timestep $t$, and let $z_{\mathrm{final}}^{i}$ denote the corresponding final denoised latent representation. We define the cosine similarity between them as
\begin{equation}
S_t^{i} 
= \frac{z_t^{i} \cdot z_{\mathrm{final}}^{i}}{\|z_t^{i}\| \, \|z_{\mathrm{final}}^{i}\|}.
\end{equation}
Here, $S_t^{i}$ serves as a practical proxy for timestep-wise alignment toward the terminal denoised state. Rather than relying on the absolute similarity value itself, we focus on how this alignment changes across consecutive timesteps, since a larger increase in similarity indicates that the corresponding transition makes a stronger relative contribution toward the final denoised outcome. Specifically, we compute the similarity difference between consecutive timesteps:
\begin{equation}
\Delta S_t^{i} = S_t^{i} - S_{t+1}^{i}.
\end{equation}
A larger $\Delta S_t^{i}$ indicates that timestep $t$ is associated with a larger increase in alignment toward the final denoised state, and is therefore treated as more important in the trajectory.

To maintain stability and compatibility with GRPO, we convert these step-wise alignment changes into normalized non-negative temporal weights:
\begin{equation}
w_t^{i} = 
\frac{\max\big(0, \Delta S_t^{i}\big) + \epsilon}
{\sum_{\tau} \left(\max(0, \Delta S_\tau^{i}) + \epsilon\right)},
\end{equation}
where $\epsilon$ is a small constant used to avoid degenerate zero weights. These weights redistribute the trajectory-level advantage across timesteps according to their relative contribution, allowing GRPO to place stronger policy updates on more informative denoising steps instead of uniformly broadcasting the same optimization signal throughout the trajectory.

\subsection{Multi-Objective Credit Allocation}
Temporal credit assignment identifies which denoising steps deserve stronger policy updates, but effective supervision in visual generation also depends on how reward signals are composed at those steps. In practice, policy optimization is driven by multiple heterogeneous objectives, such as visual quality, motion quality, and text alignment, whose interactions can be nontrivial and whose relevance may vary across the denoising trajectory. However, existing GRPO methods typically reduce these objectives to a single scalar reward and apply the resulting supervision uniformly during optimization. This fixed reward composition is often too coarse to reflect the evolving needs of generation, motivating an adaptive objective credit mechanism that assigns step-conditioned reward weights throughout the trajectory.

\vspace{1mm}
\noindent\textbf{Advantage-Space Multi-Objective Fusion.}
To address this, we formulate the balancing of competing rewards as a Multi-Objective Optimization (MOO) problem. Traditional gradient-based MOO seeks a conflict-aware, Pareto-stationary update direction by solving a minimum-norm problem:
\begin{equation}
\min_{\{c_k\}_{k=1}^{K}}\Bigl\|\sum_{k=1}^{K}c_k\nabla_{\!\theta}\mathcal{L}_k(\theta)\Bigr\|^{2}_2\quad\mathrm{s.t.}\quad \sum_{k=1}^{K} c_k=1,\;\;c_k\geq 0\;\;\forall k,
\end{equation}
where $c_k$ is the dynamic weight assigned to the $k$-th objective, and $\nabla_{\!\theta}\mathcal{L}_k(\theta)$ is the corresponding gradient. However, computing and storing per-objective gradients for modern large-scale generative models is computationally prohibitive, making this direct approach infeasible for online RL. Inspired by MGDA-UB~\cite{sener2018multi}, which proves that optimizing an upper bound of the multi-objective gradient norm yields a Pareto-stationary solution, we bypass this bottleneck by projecting the optimization from the gradient space into the advantage space. Using the chain rule from the policy gradient theorem, the gradient norm can be bounded as follows:
\begin{equation}
\begin{aligned}
\Bigl\|\sum_{k=1}^{K}c_k\nabla_{\!\theta}\mathcal{L}_k(\theta)\Bigr\|^{2}_2
&= \Bigl\|\sum_{k=1}^{K}c_k\nabla_{\!\theta}\left(\frac{\pi_\theta}{\pi_{\theta_{\text{old}}}}A_k\right)\Bigr\|^{2}_2 \\
&= \Bigl\|\sum_{k=1}^{K}c_k\frac{1}{\pi_{\theta_{\text{old}}}}A_k\nabla_{\!\theta}{\pi_\theta}\Bigr\|^{2}_2 \\
&\leq \Bigl\|\sum_{k=1}^{K}c_k A_k\Bigr\|^{2}_2\Bigl\|\frac{1}{\pi_{\theta_{\text{old}}}}\nabla_{\!\theta}{\pi_\theta}\Bigr\|^{2}_2.
\end{aligned}
\end{equation}
Since the policy gradient term $\frac{1}{\pi_{\theta_{\text{old}}}}\nabla_{\!\theta}{\pi_\theta}$ is shared across objectives for a fixed sample $i$ during a given update step, the per-objective gradients differ only by their scalar advantages $A_k^i$. Under this shared-direction structure, the gradient-space multi-objective optimization problem can be reformulated as a low-dimensional optimization over scalar advantages:
\begin{equation}
\min_{\{c_k^i\}_{k=1}^{K}}\Bigl\|\sum_{k=1}^{K}c_k^i A_k^{i}\Bigr\|^{2}\quad\mathrm{s.t.}\quad \sum_{k=1}^{K} c_k^i=1,\;\;c_k^i\geq 0\;\;\forall k,
\label{eq:adv_moo}
\end{equation}
where $A_k^{i}$ is the group-wise normalized advantage of sample $i$ for the $k$-th objective.

\vspace{1mm}
\noindent\textbf{Adaptive Exploration Mechanism.}
However, a naive application of this minimum-norm solver introduces a new challenge: \textbf{overly conservative policy updates}. Because the solver strictly penalizes variance among competing objectives, it naturally favors "safe", compromised directions where all advantages are small but equal. This cautious behavior discourages the policy from taking bold steps toward high-reward but conflicting directions, potentially trapping the generative model in mediocre local optima. 
To break this conservative deadlock and ensure robust policy discovery, we introduce a dynamic exploration mechanism. We first quantify an exploration signal $e_i$ based on the structural deviation of the trajectory from the group average:
\begin{equation}
q_i = \sum_{t=1}^{T} w_t^{i} \cdot \Delta S_t^{i}, \quad
e_i = \frac{|q_i|}{\mathrm{std}(\{q_j\}_{j=1}^{G}) + \epsilon},
\end{equation}
where $G$ is the group size. A high $e_i$ indicates that the generation path is highly distinctive, signaling a valuable region for exploration. We then inject an exploration bias into the objective-level solver:
\begin{equation}
\min_{\{c_{k}^{i}\}} \Bigl\|\sum_{k=1}^{K} c_{k}^{i} A_{k}^{i}\Bigr\|^{2} - \lambda_i \sum_{k=1}^{K} c_{k}^{i} A_{k}^{i} \quad\mathrm{s.t.}\quad \sum_{k=1}^{K} c_{k}^{i} = 1,\; c_{k}^{i} \geq 0\;\forall k,
\end{equation}
where the added penalty term encourages the solver to maximize the total weighted advantage rather than strictly minimizing the norm. To prevent unstable divergence, the exploration strength $\lambda_i$ is dynamically modulated:
\begin{equation}
\lambda_i = \frac{\max\left(0, \sum_{k=1}^{K} A_{k}^{i}\right)}{\left|\sum_{k=1}^{K} A_{k}^{i}\right| + \epsilon} \cdot \tanh(e_i).
\end{equation}
Under this formulation, exploration emerges only when the aggregated advantage is positive and the trajectory exhibits significant structural deviation. With the temporal weights $w_t^{i}$ and the adaptive objective coefficients $c_k^{i}$, the final effective advantage at timestep $t$ is defined as
\[
\tilde{A}_{t}^{i} = w_t^{i} \sum_{k=1}^{K} c_{k}^{i} A_{k}^{i}.
\]
This effective advantage provides a unified training signal for visual GRPO: it redistributes policy credit across denoising steps according to their relative contribution, while adaptively composing multiple reward signals during optimization. As a result, policy updates become better matched to the structure of the generation trajectory, yielding more stable optimization and improved generation quality.
\subsection{Optimization Target}
\label{sec:optimization}

Finally, we incorporate the proposed credit assignment scheme into the GRPO objective by replacing the original trajectory-level advantage $A_i$ in Eq.~\ref{eq:grpo} with the timestep-specific effective advantage $\tilde{A}_{t}^{i}$. The resulting policy objective is defined as:
\begin{equation}
\begin{aligned}
\mathcal{J}_{\text{OTCA}}(\theta) &= \mathbb{E}_{\substack{\{\mathbf{o}_{i}\}_{i=1}^{G}\sim\pi_{\theta_{\mathrm{old}}}(\cdot|\mathbf{c}) \\ 
\mathbf{a}_{t,i}\sim\pi_{\theta_{\mathrm{old}}}(\cdot|\mathbf{s}_{t,i})}}
\Bigg[\frac{1}{G}\sum_{i=1}^{G}\frac{1}{T}\sum_{t=1}^{T} \\
&\qquad\quad \min\Big(\rho^i_{t}\tilde{A}_{t}^{i}, \, \mathrm{clip}(\rho^i_{t},1-\epsilon,1+\epsilon)\tilde{A}_{t}^{i}\Big)\Bigg].
\end{aligned}
\end{equation}
By combining timestep-sensitive weighting with adaptive reward composition into a single effective advantage, OTCA provides a unified training signal for visual GRPO. This effective advantage allows policy updates to concentrate on more informative denoising steps while adapting the reward composition to the evolving needs of the generation trajectory.

As a result, OTCA makes reinforcement learning updates better aligned with the multi-stage nature of visual generation. Rather than uniformly broadcasting a coarse global reward across the entire trajectory, the resulting policy signal is redistributed according to step-wise contribution and reward relevance, leading to more stable optimization and improved generation quality.

\section{Experiment}
\label{sec:experiment}
\subsection{Settings}
\label{sec:settings}
\noindent\textbf{Datasets.} Both image and Video generation are finetuned with the prompts from DanceGRPO, which contains 50k prompts spanning various styles. We randomly separate 1K data as our test set for image generation.

\noindent\textbf{Backbones and Rewards.}
For image generation, we fine-tune FLUX.1-dev~\cite{blackforest2024flux} using three reward models: CLIP-T~\cite{radford2021learning}, HPSv2.1 \cite{hpsv2}, and the LAION aesthetic predictor~\cite{schuhmann2022laion}. We assess generalization on both in-domain and out-of-domain benchmarks including PickScore \cite{pickscore} and ImageReward~\cite{xu2023imagereward}. For video generation, we fine-tune Wan2.2-T2V-14B-480P~\cite{wan2025wan} with VideoAlign, which provides reward signals along three dimensions: visual quality, motion quality, and text alignment. For out-of-domain evaluation on video generation, we adopt VBench~\cite{huang2024vbench}, which evaluates video quality across multiple dimensions.

\noindent\textbf{Implementation Details.} For image generation, we set the group size to \(G = 12\), downsample training resolution to \(512 \times 512\), and use 16 sampling steps. For video generation, we set the group size to \(G = 32\), training resolution to \(240 \times 416 \times 53\) (\(H \times W \times T\)), and use 16 sampling steps to accelerate training. During inference, we increase the resolution to \(512 \times 512\) with 50 inference steps for Flux, and \(480 \times 832 \times 53\) with 50 inference steps for Wan2.2-T2V-14B. All image generation experiments are conducted on 8 NVIDIA H100 GPUs, while video generation experiments are trained on 32 NVIDIA H100 GPUs. The stochasticity parameter \(\eta\) is set to 0.3 for Flux and 0.8 for Wan2.2, with a learning rate of \(1 \times 10^{-5}\) and \(5 \times 10^{-6}\) respectively. 
\begin{figure*}[htbp]
    \centering
    \includegraphics[width=0.98\textwidth]{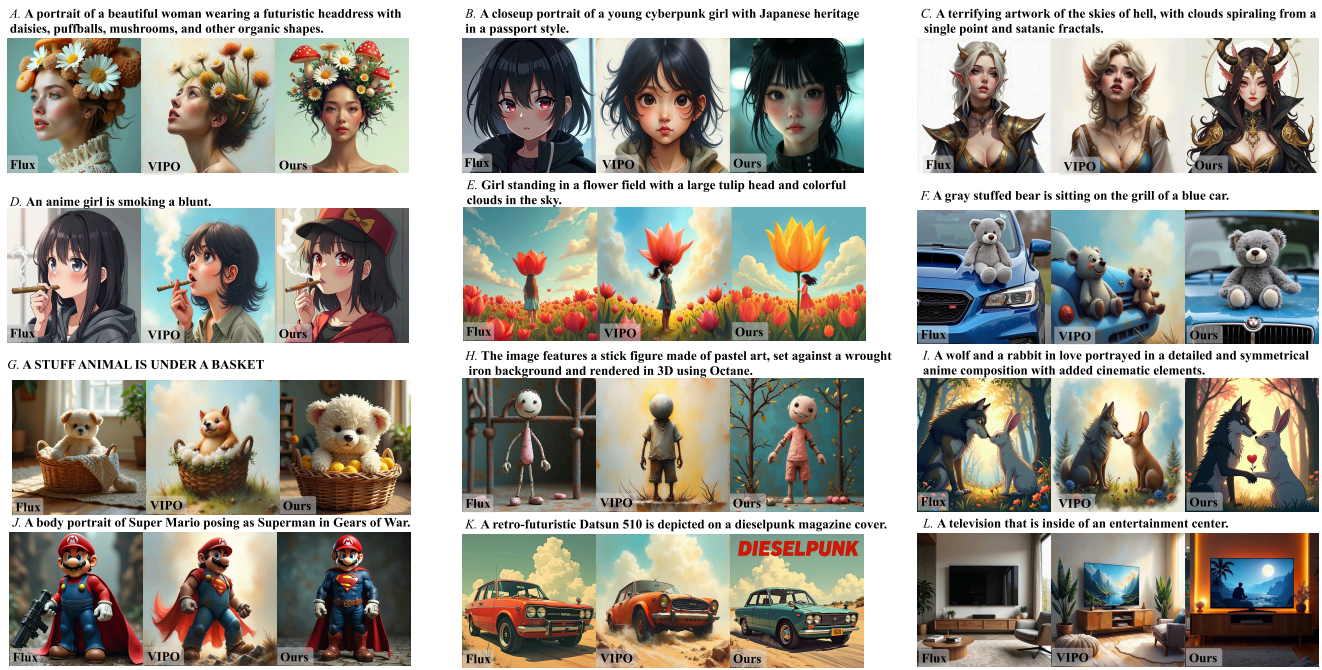}
    \caption{Qualitative comparison on image generation. Our method produces images with higher fidelity and aesthetic quality compared to existing GRPO-based approaches, demonstrating clearer structures and more consistent temporal alignment.}
\label{fig:qualitative_flux}
\end{figure*}
\subsection{Quantitative Results}
\subsubsection{Image Generation Quantitative Results.}
We evaluate OTCA on FLUX.1-dev and compare it against the base FLUX model, DanceGRPO, and VIPO. We first report the in-domain metrics used during training, including CLIP-T, LAION-Aesthetic, and HPSv2.1. To further assess whether the improvements extend beyond the training reward distribution, we additionally evaluate out-of-domain generalization using PickScore and ImageReward.

As shown in Table~\ref{tab:image_reward_comparison}, OTCA outperforms the base model and prior GRPO-based variants on most evaluation metrics. It achieves the strongest results on preference-oriented measures, including CLIP-T, HPS, PickScore, and ImageReward. The improvements are consistent across both in-domain and out-of-domain evaluations, indicating better generalization beyond the training rewards.

In contrast, DanceGRPO and VIPO show less balanced performance. For instance, while DanceGRPO yields competitive aesthetic scores, it experiences noticeable degradation on semantic-oriented metrics. This discrepancy indicates a common limitation of uniform reward broadcasting: the policy tends to disproportionately favor aesthetic quality at the expense of semantic fidelity. OTCA effectively mitigates this optimization bias via adaptive multi-objective allocation, yielding a more favorable trade-off across diverse preference dimensions, as reflected by the consistent gains in CLIP-T and PickScore. Similarly, although VIPO generally improves upon the base model, it consistently underperforms OTCA across key preference measures. These results suggest that simply combining multiple rewards is insufficient for stable alignment.

Overall, the results show that reorganizing reward supervision along the generation trajectory leads to more balanced optimization. By coordinating credit assignment across timesteps and objectives, OTCA better captures multi-dimensional human preferences.
\label{sec:image_results}
\begin{table}[t]
\centering
\caption{Comparison of different methods on image-level reward metrics.}
\label{tab:image_reward_comparison}
\setlength{\tabcolsep}{3pt}
\small
\resizebox{\columnwidth}{!}{%
\begin{tabular}{lccccc}
\toprule
Method & CLIP-T $\uparrow$ & Aesthetic $\uparrow$ & HPS $\uparrow$ & PickScore $\uparrow$ & ImageReward $\uparrow$ \\
\midrule
Flux        & 0.2682 & 6.2508 & 0.2986 & 22.66 & 1.1001 \\
DanceGRPO    & 0.2619 & \textbf{6.8917} & 0.3018 & 22.70 & 1.0172 \\
VIPO        & 0.2722 & 6.7813 & 0.3026 & 22.69 & 1.1128 \\
Ours & \textbf{0.3071} & 6.6028 & \textbf{0.3225} & \textbf{22.97} & \textbf{1.1998} \\
\bottomrule
\end{tabular}%
}
\end{table}

\subsubsection{Video Generation Quantitative Results}
We evaluate OTCA on VBench against Wan2.2 and DanceGRPO~\cite{dancegrpo}, a recent GRPO-based baseline for visual generation. Tables~\ref{tab:vbench_half} and~\ref{tab:vbench_overall} report the dimension-wise and overall results, respectively.

As shown in Table~\ref{tab:vbench_half}, OTCA improves most evaluation dimensions over both baselines. Gains in Dynamic Degree indicate better motion fidelity, while improvements in Spatial Relationship and Multiple Objects reflect stronger spatial coherence and multi-object handling. Gains in Color, Scene, and Subject Consistency further indicate improved visual fidelity and semantic alignment.

\begin{figure*}[htbp]
    \centering
    \includegraphics[width=0.98\textwidth]{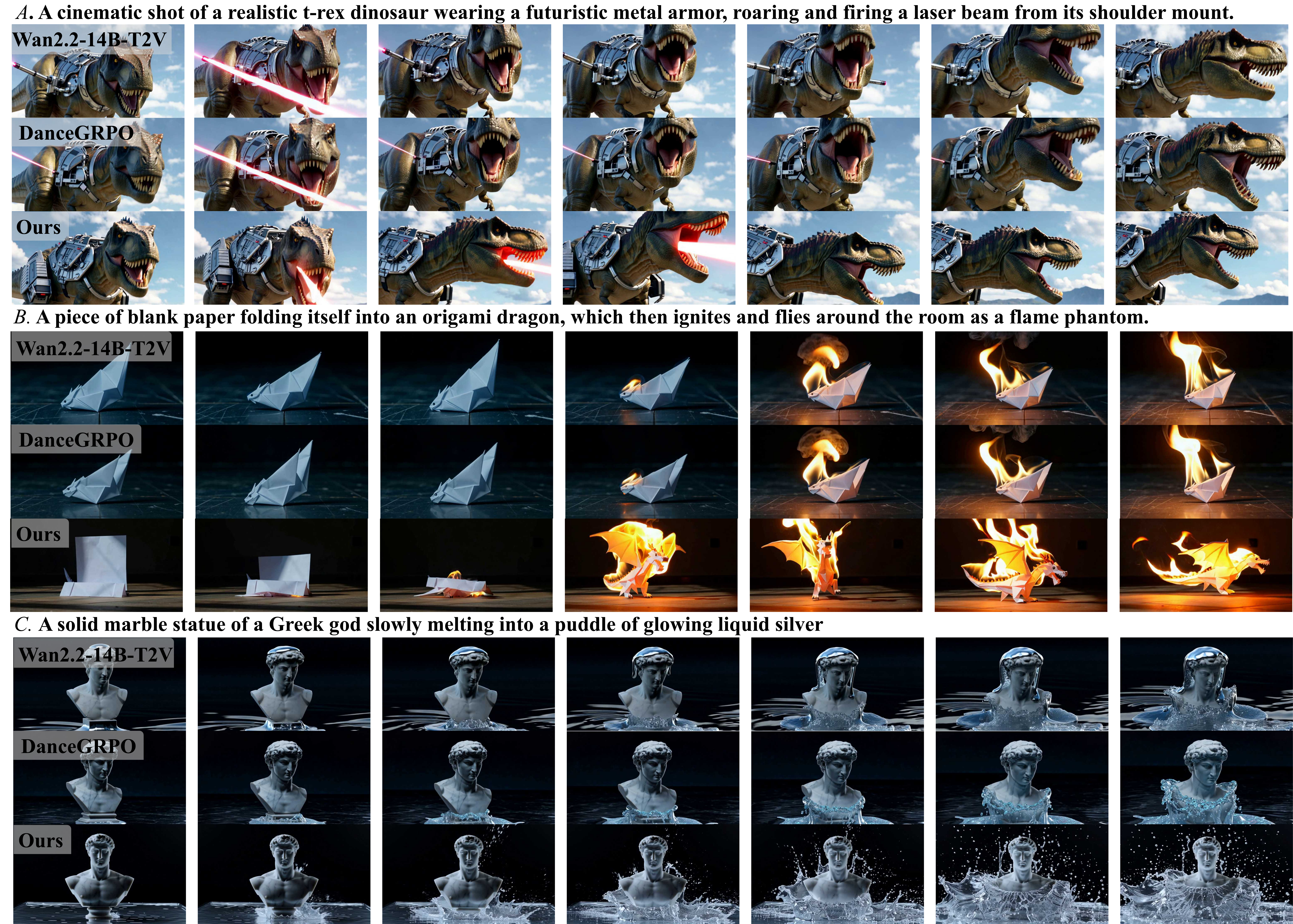}
    \caption{Qualitative comparison on video generation. From top to bottom: results generated by Wan2.2 (baseline), DanceGRPO, and our proposed OTCA. Our method produces more detailed and perceptually pleasing results, with richer textures, more accurate colors, and better semantic alignment to the prompt.}
    \label{fig:qualitative_video}
\end{figure*}

Notably, VideoAlign provides separate rewards for Visual Quality, Motion Quality, and Text Alignment. MOCA coordinates these objectives, while TCD assigns stronger supervision to timesteps that contribute more to the final output. Their joint objective- and timestep-level credit assignment improves global composition and semantic structure, as reflected by gains in Scene and Spatial Relationship. OTCA also achieves higher overall Quality, Semantic, and Total Scores than both baselines, establishing it as a strong baseline for video generation post-training.

We attribute these improvements to two complementary factors. First, MOCA coordinates heterogeneous rewards, reducing conflicts that are particularly pronounced in video generation, where SDE-based exploration often yields low-quality samples and video reward models remain less precise than their image counterparts. Second, TCD redistributes supervision across denoising timesteps rather than broadcasting a uniform signal, accounting for their unequal contributions to the final output. Together, these mechanisms improve objective compatibility and temporal precision, leading to more stable and effective training.

\subsection{Qualitative Experiment}

\begin{table}[t]
\centering
\caption{Quantitative comparison on the VBench benchmark. OTCA consistently outperforms other methods across key dimensions.}
\label{tab:vbench_half}
\setlength{\tabcolsep}{2.2pt}
\footnotesize
\resizebox{\linewidth}{!}{%
\begin{tabular}{lccccccc}
\toprule
Method & Color & Dynamic & Imaging & Multiple & Scene & Spatial & Subject \\
       &        & Degree  & Quality & Objects  &       & Relation & Consistency \\
\midrule
Wan2.2       & 89.79 & 70.83 & 68.34 & 63.79 & 35.90 & 80.54 & 93.78 \\
DanceGRPO    & 90.35 & 71.45 & 67.93 & 64.85 & 35.18 & 83.25 & 93.76 \\
OTCA (Ours)  & \textbf{92.88} & \textbf{75.00} & \textbf{68.58} & \textbf{68.45} & \textbf{38.45} & \textbf{88.57} & \textbf{93.95} \\
\bottomrule
\end{tabular}%
}
\end{table}

\begin{table}[t]
\centering
\caption{Overall VBench scores comparison. OTCA achieves consistent improvements across quality, semantic, and total scores.}
\label{tab:vbench_overall}
\resizebox{\linewidth}{!}{%
\begin{tabular}{lccc}
\toprule
\textbf{Method} & \textbf{Quality Score} & \textbf{Semantic Score} & \textbf{Total Score} \\
\midrule
Wan2.2        & 83.30 & 73.10 & 81.26 \\
DanceGRPO     & 83.31 & 73.31 & 81.31 \\
Ours   & \textbf{83.72} & \textbf{75.19} & \textbf{82.01} \\
\bottomrule
\end{tabular}%
}
\end{table}
\subsubsection{Image Generation Qualitative Results}
As shown in Fig.~\ref{fig:qualitative_flux}, OTCA produces more faithful and visually appealing images than the baselines. For human-related prompts, it generates more coherent layouts, richer details, and stronger realism, while for object-centric prompts, it yields more vivid anthropomorphic effects and more diverse visual expressions. In challenging cases like case H and I, OTCA better captures the interactions and preserves semantic consistency, whereas even the strong baseline VIPO may degrade, likely due to its limited coordination across multiple reward signals and the lack of fine-grained timestep-level credit assignment.
\subsubsection{Video Generation Qualitative Results}
As shown in Fig.~\ref{fig:qualitative_video}, in Case A, Wan2.2 and DanceGRPO produce implausible dragon breath spanning the entire frame, whereas OTCA generates a realistic, localized effect. In Case B, neither baseline captures a sheet of paper folding into a dragon, while OTCA successfully renders the transformation with better visual quality. In Case C, OTCA produces more realistic liquid dynamics.

These qualitative examples provide further evidence for the effectiveness of OTCA. OTCA generates videos with better physical plausibility, more faithful motion evolution, and stronger semantic consistency. We attribute these gains to adaptively coordinating heterogeneous rewards and redistributing supervision across denoising timesteps.

\subsection{Ablation Studies}
\label{sec:ablation}

\begin{table}[t]
\centering
\caption{Ablation study on image-level reward metrics. Best results are highlighted in bold.}
\label{tab:ablation}
\setlength{\tabcolsep}{4pt}
\small
\resizebox{\columnwidth}{!}{%
\begin{tabular}{lccccc}
\toprule
Ablation & CLIP-T $\uparrow$ & Aesthetic $\uparrow$ & HPS $\uparrow$ & PickScore $\uparrow$ & ImageReward $\uparrow$ \\
\midrule
Flux         & 0.2682 & 6.2508 & 0.2986 & 22.66 & 1.1001 \\
MOCA-only    & 0.2787 & 6.2120 & 0.3049  & 22.64 & 1.1250 \\
TCD-only & 0.2708 & 6.3270 & 0.3104 & 22.87 & \textbf{1.2618} \\
Ours (Full)        & \textbf{0.3071} & \textbf{6.6028} & \textbf{0.3225} & \textbf{22.97} & 1.1998 \\
\bottomrule
\end{tabular}%
}
\end{table} 

\subsubsection{Reward Metric Results}
We conduct an ablation study on image-level reward metrics to isolate the contributions of TCD and MOCA. Table~\ref{tab:ablation} compares the full model with \emph{TCD-only} and \emph{MOCA-only}. Both variants improve several metrics over FLUX, confirming the effectiveness of each component.



Specifically, \emph{TCD-Only} improves CLIP-T and HPS, indicating that trajectory-level credit decomposition better allocates supervision across denoising steps. Its limited aesthetic gain, however, suggests that temporal credit alone cannot fully capture high-level preferences without multi-reward integration.

In contrast, \emph{MOCA-Only} improves HPS and ImageReward, showing that direct reward fusion strengthens reward-oriented alignment. However, it remains weaker than the full model on the other metrics, suggesting that reward integration alone may favor specific objectives rather than improve overall perceptual quality.

Our full model achieves the best CLIP-T, Aesthetic, HPS, and PickScore and the second-best ImageReward. These results confirm the complementarity of TCD and MOCA: TCD determines where supervision should be emphasized, whereas MOCA determines what should be optimized. Their combination yields the most balanced overall performance.

\subsubsection{Reward Curve Results}
\begin{figure}[!t]
    \centering
    \includegraphics[width=\columnwidth]{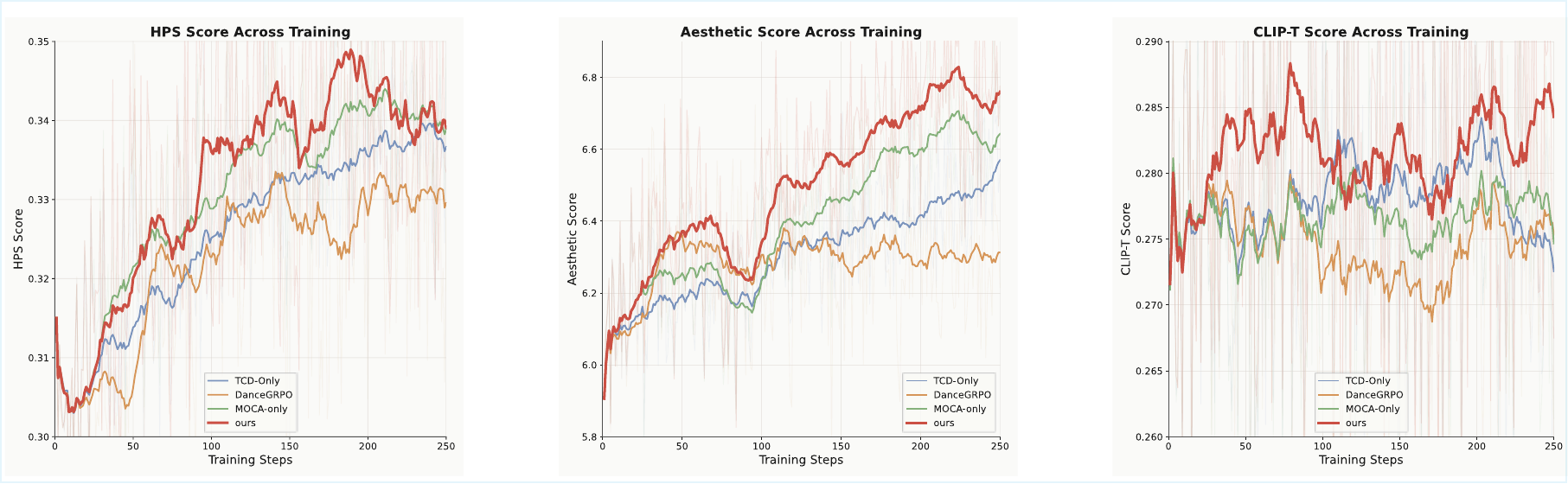}
    \caption{Reward curves on different ablation during training.}
    \label{fig:reward_flux}
\end{figure}

To further examine the contribution of different design choices, we present the reward curves of all variants throughout training. As shown in the figure~\ref{fig:reward_flux}, the full model converges faster and achieves higher final reward values across all three metrics than the simplified variants. Using only the latent credit assign or only the reward signal integration already yields clear improvements over the baseline, while their combination delivers the strongest overall performance, indicating that the two signals provide complementary supervision. Moreover, compared with DanceGRPO, our method exhibits a more stable upward trend during training. The curve experiment suggest that OTCA can effectively integrate multi‑source reward signals into each denoising timestep, providing more reliable and effective optimization.

\section{Conclusion}
In this paper, we present OTCA, an structured reward credit modeling framework for GRPO-based visual generation. Our key insight is that effective optimization requires answering two coupled questions: which timesteps deserve stronger policy supervision, and which reward objectives should dominate at each stage of generation. To this end, OTCA introduces a unified credit assignment mechanism that jointly models timestep importance and adaptive multi-objective fusion, transforming coarse trajectory-level supervision into fine-grained, process-aware policy updates. Extensive experiments on both image and video generation demonstrate that OTCA consistently improves visual quality, semantic alignment, and optimization stability over strong GRPO-based baselines. Ablation studies further verify that temporal credit assignment and adaptive reward integration contribute complementary benefits. 

\bibliographystyle{ACM-Reference-Format}
\bibliography{main}

\appendix

\section{Appendix}
\subsection{Timestep Contribution Proxy}
A key assumption of our method is that $\Delta S_t^i$ serves as a reliable proxy for the contribution of each denoising timestep. To examine this assumption, we decode the intermediate latent state $Z_t^i$ at each timestep $t$ along the sampling trajectory and compare $\Delta S_t^i$ with the corresponding reward improvement $\Delta R_t^i$. Intuitively, if $\Delta S_t^i$ faithfully captures timestep-wise contribution to the final generation outcome, then timesteps with larger $\Delta S_t^i$ should also be associated with larger reward gains, respectively larger contribution.

\textbf{Quantitative evidence.}
Table~\ref{tab:delta-sim-delta-reward} summarizes the quantitative relationship between $\Delta S_t^i$ and $\Delta R_t^i$ over 511 sampled trajectories. First, we examine their global correlation by analysing the correlation coefficient. The results indicates that $\Delta S_t^i$ exhibits a strong positive linear association with $\Delta R_t^i$. In detail, the Pearson correlation achieves $0.909$, highly correlated. This indicates that timesteps with larger latent similarity increase tend to gain larger reward improvements. The same trend also exsits at the ranking level, the Spearman correlation reaches $0.838$, suggesting that $\Delta S_t^i$ remains highly consistent with $\Delta R_t^i$ in capturing the relative importance of different denoising steps.

Beyond global correlation, we further assess whether $\Delta S_t^i$ can accurately localize the timesteps that contribute most to reward improvement. The pairwise order agreement, which measures the fraction of timestep pairs whose relative ordering is consistent between $\Delta S_t^i$ and the true reward change $\Delta R_t^i$, reaches $0.842$. This shows that the ranking induced by $\Delta S_t^i$ closely matches that of $\Delta R_t^i$. In addition, the average argmax distance between the peaks of $\Delta S_t^i$ and $\Delta R_t^i$ is only $0.74$ steps in the total $16$ sampling steps, indicating that the most salient timestep identified by $\Delta S_t^i$ is typically very close to the true reward-critical step. This localization ability is further supported by the substantial overlap between the high-response regions, with $\mathrm{Recall@3}=0.770$ and $\mathrm{Recall@5}=0.892$.

Overall, these results consistently support the use of $\Delta S_t^i$ as an effective proxy for timestep contribution. These metrics provide empirical justification for using $\Delta S_t^i$ to allocate fine-grained optimization strength across denoising steps.

\begin{table}[t]
\centering
\caption{Correlation analysis between $\Delta{S^i_t}$ and $\Delta{R^i_t}$. The results support the use of $\Delta{S^i_t}$ as a proxy for timestep contribution.}
\label{tab:delta-sim-delta-reward}
\begin{tabular}{lcc}
\toprule
Metric & Range & Value \\
\midrule
Pearson correlation $\uparrow$ & $[-1, 1]$ & $0.9091 \pm 0.028$ \\
Spearman correlation $\uparrow$ & $[-1, 1]$ & $0.8381 \pm 0.077$ \\
Pairwise order agreement $\uparrow$ & $[0, 1]$ & $0.8428 \pm 0.049$ \\
Recall@3 $\uparrow$ & $[0, 1]$ & $0.770$ \\
Recall@5 $\uparrow$ & $[0, 1]$ & $0.892$ \\
Argmax distance $\downarrow$ & $[0, 15]$ & $0.74$ \\
\bottomrule
\end{tabular}
\end{table}

\begin{figure*}[h]
    \centering
    \includegraphics[width=\textwidth]{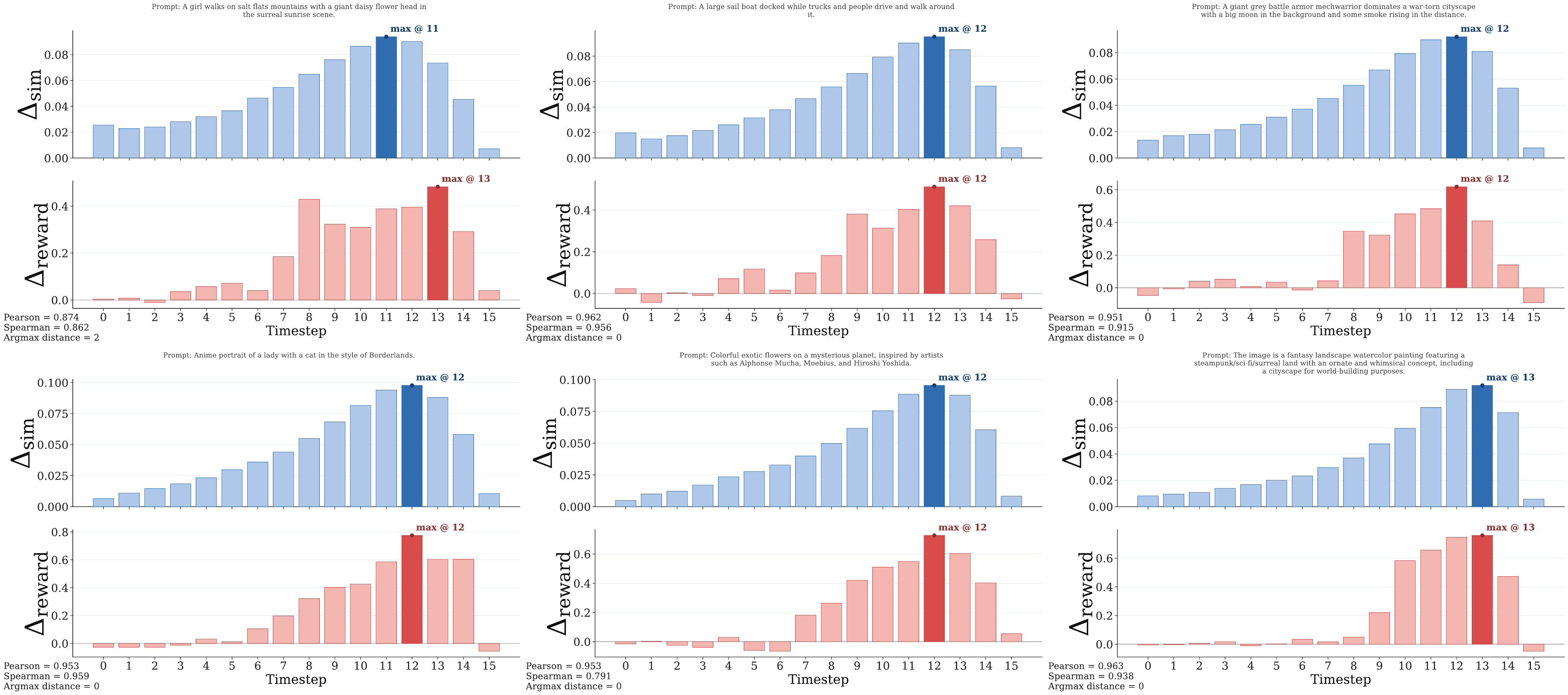}
    \caption{Qualitative evidence supporting the use of $\Delta S^i_t$  as a proxy for timestep contribution. The figure shows a clear correlation between $\Delta S^i_t$ and $\Delta R^i_t$.}
    \label{fig:timestepgan_flux}
\end{figure*}
\textbf{Qualitative evidence.}
To further provide intuitive evidence for the validity of $\Delta S_t^i$ as a proxy for timestep contribution, we visualize the relationship between $\Delta S_t^i$ and $\Delta R_t^i$ along the sampling trajectory for several representative examples in Fig.~\ref{fig:timestepgan_flux}. As shown, the two signals exhibit highly consistent temporal patterns across different cases. Timesteps with larger $\Delta S_t^i$ generally coincide with more pronounced reward improvements, while low-response steps tend to yield only limited reward gains. This suggests that $\Delta S_t^i$ captures the relative contribution of each timestep to the final quality improvement.

Moreover, as shown in Fig.~\ref{fig:timestepgan_flux}, in most illustrated cases the timestep with the largest $\Delta S_t^i$ tends to yield the greatest reward gain $\Delta R_t^i$. Even when not perfectly aligned, the deviation is typically limited to only 1--2 neighboring timesteps and still falls within the same high-response region. The above qualitative experiments further demonstrate the reliability of using timestep increase as an indicator of each timestep contribution.

Overall, these qualitative observations are highly consistent with the quantitative findings.
From an intuitive perspective, $\Delta S_t^i$ closely tracks the evolution of reward improvement along the denoising trajectory and effectively localizes the denoising steps that are most critical to the final reward gain, further supporting its use as an effective timestep-wise contribution proxy.

\subsection{Exploration-Biased Objective Formulation}

We briefly analyze the exploration bias introduced in Eq.~(12)–(14). 
The trajectory score $q_i = \sum_t w_t^i \Delta S_t^i$ summarizes the weighted structural contribution of a sampled path across denoising steps. A larger $|q_i|$ indicates that the trajectory exhibits stronger step-wise alignment changes toward the final state, and therefore reflects a more distinctive and potentially informative generation process. Such trajectories deviate more noticeably from the group-average evolution pattern and are thus considered more worthwhile to explore.

To make this signal comparable across samples within a group, we normalize it using the group standard deviation, yielding 
$e_i = |q_i| / (\mathrm{std}(\{q_j\}) + \epsilon)$. 
This standardization prevents the exploration strength from being dominated by absolute scale variations across prompts, ensuring that the bias reflects relative structural deviation rather than raw magnitude.

The exploration bias is activated only when the aggregated advantage is positive. This constraint ensures that we encourage directions that are already reward-consistent, rather than amplifying undesirable updates. The modulation term $\tanh(e_i)$ further keeps the exploration strength smoothly bounded, avoiding abrupt or unstable perturbations while preserving monotonic sensitivity to structural deviation.

With the introduction of $\lambda_i$, the original minimum-norm objective no longer purely favors the smallest aggregated advantage. Instead, it explicitly trades off between minimizing disagreement and retaining larger positive advantages. As a result, objective mixtures associated with stronger reward signals receive increased preference, while the coefficient vector remains constrained on the probability simplex. When $\lambda_i = 0$, Eq.~(13) reduces to the original minimum-norm objective in Eq.~(11), and the solver degenerates to the conservative minimum-norm fusion. Therefore, the proposed exploration bias can be viewed as a smooth extension rather than a structural modification of the original formulation.

The multi-objective optimization in Eq.~(13) can be equivalently reduced to a one-dimensional quadratic problem over the aggregated scalar $z = \mathbf{c}^{i\top}\mathbf{A}^i$:
\[
\min_{z \in [s_{\min}, s_{\max}]} z^2 - \lambda_i z,
\]
where $[s_{\min}, s_{\max}]$ denotes the feasible range induced by individual advantages. 
The objective $z^2 - \lambda_i z$ is a convex quadratic function in $z$, ensuring a unique global minimizer prior to projection. 
The solver in Algorithm~\ref{alg:moca_solver} first computes the unconstrained minimizer $z^* = \lambda_i/2$, projects it onto this interval, and then recovers a coefficient vector via linear interpolation between neighboring objectives. This closed-form procedure is efficient, preserves the simplex constraint, and introduces only a smooth perturbation to the original minimum-norm formulation.

\subsection{Algorithm Presentation.}
To facilitate a clearer understanding of our method, we provide pseudocode-style frameworks of the proposed algorithms. Specifically, Algorithm~\ref{alg:moca_solver} details the integration process among multiple reward components in the MOCA module, while Algorithm~\ref{alg:otca_overall} illustrates the overall optimization pipeline of OTCA. These algorithmic presentations are intended to highlight the logical flow and design principles, rather than executable code.

\clearpage
\begin{algorithm}[H]
\caption{Multi-Objective Credit Allocation (MOCA) Solver}
\label{alg:moca_solver}
\begin{algorithmic}[1]
\Require Advantages $\mathbf{A}^i = [A_1^i, \dots, A_K^i]$ for sample $i$;
         exploration strength $\lambda_i$; tolerance $\epsilon = 10^{-8}$
\Ensure  Coefficient vector $\mathbf{c}^i \in \Delta^K$ (probability simplex)

\vspace{4pt}
\State \textbf{// Step 1: Determine feasible range of $z = \mathbf{c}^{i\top}\mathbf{A}^i$}
\State $s_{\min} \leftarrow \min_k A_k^i$, \quad $s_{\max} \leftarrow \max_k A_k^i$

\vspace{4pt}
\If{$s_{\max} - s_{\min} < \epsilon$}                       \Comment{Case 1: all advantages equal}
    \State $k^* \leftarrow \arg\max_k A_k^i$
    \State \Return $\mathbf{c}^i$ with $c^i_{k^*} = 1$ and $c^i_k = 0\ \forall k \neq k^*$
\EndIf

\vspace{4pt}
\State \textbf{// Step 2: Closed-form solution of $\min_{z}\, z^2 - \lambda_i z$, projected onto $[s_{\min}, s_{\max}]$}
\State $z^* \leftarrow \lambda_i / 2$    
\State $\hat{z} \leftarrow \mathrm{clip}(z^*,\; s_{\min},\; s_{\max})$

\vspace{4pt}
\If{$\exists\, k \text{ s.t. } |A_k^i - \hat{z}| < \epsilon$}   \Comment{Case 2: $\hat{z}$ matches $A_k^i$ exactly}
    \State \Return $\mathbf{c}^i$ with $c^i_k = 1$ and $c^i_{k'} = 0\ \forall k' \neq k$
\EndIf

\vspace{4pt}
\State \textbf{// Step 3: Recover sparse $\mathbf{c}^i$ via linear interpolation}
\State Sort indices so that $A_{(1)} \leq \cdots \leq A_{(K)}$, with permutation $\pi$
\State $\mathbf{c}^i \leftarrow \mathbf{0}$
\For{$j = 1, \dots, K-1$}
    \If{$A_{(j)} \leq \hat{z} \leq A_{(j+1)}$} \Comment{Case 3: bracket found}
        \State $c^i_{\pi(j)}   \leftarrow \dfrac{A_{(j+1)} - \hat{z}}{A_{(j+1)} - A_{(j)}}$
        \State $c^i_{\pi(j+1)} \leftarrow \dfrac{\hat{z} - A_{(j)}}{A_{(j+1)} - A_{(j)}}$
        
        \State \textbf{break}
    \EndIf
\EndFor
\State \Return $\mathbf{c}^i$
\end{algorithmic}
\end{algorithm}

\begin{algorithm}[H]
\caption{Objective-Aware Trajectory Credit Assignment (OTCA)}
\label{alg:otca_overall}
\begin{algorithmic}[1]
\Require Advantages $\mathbf{A}^i$;
         latent trajectory $\{z_t^i\}_{t=1}^T$;
         final state $z_{\text{final}}^i$;
         group statistics $\{q_j\}_{j=1}^G$;
         tolerance $\epsilon$
\Ensure Timestep-specific effective advantages $\{\tilde A_t^i\}$
\vspace{4pt}
\State \textbf{// Step 1: Trajectory-Level Credit Decomposition (TCD)}
\State Compute similarity $S_t^i$ between $z_t^i$ and $z_{\text{final}}^i$
\State $\Delta S_t^i \leftarrow S_t^i - S_{t+1}^i$
\State $w_t^i \leftarrow \mathrm{Normalize}\big(\max(0, \Delta S_t^i)\big)$
\vspace{4pt}
\State \textbf{// Step 2: Multi-Objective Credit Allocation (MOCA)}
\State $q_i \leftarrow \sum_t w_t^i \Delta S_t^i$
\State $e_i \leftarrow \dfrac{|q_i|}{\mathrm{std}_j(q_j) + \epsilon}$
\State $\lambda_i \leftarrow 
\dfrac{\max(0, \sum_k A_k^i)}
{|\sum_k A_k^i| + \epsilon}
\cdot \tanh(e_i)$
\State $\mathbf{c}^i \leftarrow 
\textsc{MOCASolver}(\mathbf{A}^i, \lambda_i)$
\Comment{Algorithm~\ref{alg:moca_solver}}
\vspace{4pt}
\State \textbf{// Step 3: Effective Advantage Construction}
\State $\tilde A_t^i \leftarrow w_t^i \cdot \mathbf{c}^{i\top}\mathbf{A}^i$
\State \Return $\{\tilde A_t^i\}$
\end{algorithmic}
\end{algorithm}

\subsection{Additional Qualitative Experiments}

To further validate the effectiveness of OTCA, we provide additional qualitative results for both image and video generation. As shown in Fig.~\ref{fig:fluxcase}, Fig.~\ref{fig:wancase1}, and Fig.~\ref{fig:wancase2}, OTCA consistently produces more coherent and better-aligned outputs across diverse prompts and visual scenarios. These additional examples provide intuitive evidence that structured, timestep-aware reward credit assignment improves generation quality in both image and video domains.

\subsubsection{Image Generation}

\begin{figure*}[t]
    \centering
    \includegraphics[width=\textwidth]{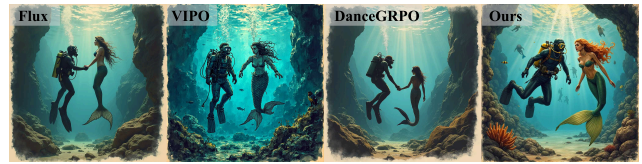}
    \caption{\textbf{More qualitative comparison of image generation.} For each text prompt, the generated videos are shown from top to bottom using Flux1.0-dev, VIPO, DanceGRPO, and our proposed method, respectively.}
    \label{fig:fluxcase}
\end{figure*}

Figure~\ref{fig:fluxcase} presents additional qualitative results on image generation. Compared with the baselines, OTCA produces images with clearer subject structure, more faithful semantic correspondence to the prompt, and richer local details. These improvements are particularly evident in challenging cases that require coordinated optimization over multiple visual aspects, where coarse global supervision often leads to under-refined regions or inconsistent object appearance. Such results reflect the benefit of Trajectory-Level Credit Decomposition (TCD), which identifies the denoising steps that are most critical to the final output and enables timestep-specific optimization rather than uniformly broadcasting the same reward signal across the entire trajectory.

\subsubsection{Video Generation}

Figures~\ref{fig:wancase1} and \ref{fig:wancase2} show additional qualitative results on video generation. OTCA yields videos with stronger semantic alignment, more coherent motion dynamics, and improved visual consistency across frames. In complex prompts involving multiple objectives, OTCA better preserves both appearance quality and temporal plausibility, while baseline methods are more prone to structural inconsistency or insufficient refinement. These gains highlight the importance of Multi-Objective Credit Allocation (MOCA), which adaptively balances heterogeneous reward signals in the advantage space, as well as TCD, which focuses optimization on the most reward-sensitive denoising stages. Together, these two components allow OTCA to refine both \emph{when} to optimize and \emph{what} to optimize, leading to more stable and effective visual GRPO training.

Overall, these additional qualitative examples further support our central claim that OTCA successfully transforms coarse global reward supervision into fine-grained, timestep-aware policy updates, resulting in stronger visual quality and more reliable alignment across both image and video generation.

\subsection{Implementation Details}
For TCD module, we use the weighting strategy with minimum weight $w_{\min}=0.5$. The remaining timesteps are correspondingly upweighted to promote trajectory diversity and mitigate reward sparsity. Additionally, no timestep fraction is applied, which means all timesteps participate in optimization. For the exploratory MOCA module, we set $\epsilon=10^{-6}$. We similarly rescale individual reward components to increase differentiation across reward terms and alleviate sparsity. For GRPO optimization, the importance ratio $\rho_t^i$ is clipped at $10^{-4}$. During sampling, we generate 12 samples per prompt for image generation and 8 for video generation. We set the gradient accumulation steps to $12$ for image generation  and $8$ for video generation and use a learning rate of $1\times10^{-5}$. Same noise for generation is adopted.

\begin{figure*}[t]
    \centering
    \includegraphics[width=0.95\textwidth]{figs/sul_wan_case.pdf}
    \caption{\textbf{More qualitative comparison of video generation.} For each text prompt, the generated videos are shown from top to bottom using Wan2.2, DanceGRPO, and our proposed method, respectively.}
    \label{fig:wancase1}
\end{figure*}
\begin{figure*}[t]
    \centering
    \includegraphics[width=0.95\textwidth]{figs/sul_wan_case1.pdf}
    \caption{\textbf{More qualitative comparison of video generation.} For each text prompt, the generated videos are shown from top to bottom using Wan2.2, DanceGRPO, and our proposed method, respectively.}
    \label{fig:wancase2}
\end{figure*}

\end{document}